\documentclass[journal]{IEEEtran}

\usepackage{amsmath,amssymb,amsfonts}
\usepackage{graphicx}
\graphicspath{{../}{./}}
\usepackage{booktabs}
\usepackage{array}
\usepackage{xcolor}
\usepackage[hidelinks]{hyperref}
\usepackage{url}

\newif\ifshowchanges \showchangestrue

\newcommand{\embLlamaEightB}{$0.91$\,GWh}      
\newcommand{\reasoningBlow}{$\sim\!18\times$}  
\newcommand{\jpertokSeventyBthru}{0.45}        
\newcommand{\jpertokSeventyBmeas}{0.39}        
\newcommand{\multiagentx}{$\sim\!15\times$}    
\newcommand{\crossoverG}{\ensuremath{1.2\times10^{10}}}   

\newcommand{\jpertokEightBthru}{0.025}         
\newcommand{\jpertokEightBmeas}{0.07}          
\newcommand{\jpertokQwenthru}{0.021}           
\newcommand{\jpertokQwenmeas}{0.07}            

\newcommand{\agecal}{\textit{agentic-eCAL}}
\newcommand{\ecal}{\textit{eCAL}}

\begin{document}



\title{Where Should Agents Live? \\Energy-Memory Characterization of Agentic AI for the Edge-Cloud Continuum}

\author{Carolina~Fortuna, Vid~Han\v{z}el, Tim~Strnad and~Bla\v{z}~Bertalani\v{c}%
\thanks{The authors are with the Department of Communication Systems, Jo\v{z}ef Stefan Institute, Jamova cesta 39, 1000 Ljubljana, Slovenia (e-mail: \{carolina.fortuna, blaz.bertalanic\}@ijs.si). Manuscript submitted to the IEEE Journal on Selected Areas in Communications (JSAC) Special Issue on Agentic AI for Intelligent Networks.}}

\maketitle

\begin{abstract}
As telecommunication networks evolve toward autonomous 5G-Advanced and 6G operations, agentic artificial intelligence (AI) workflows, where large language models (LLMs) execute multi-step reasoning, invoke diagnostic tools, retrieve domain knowledge, and coordinate across agent teams, are increasingly embedded across the edge-cloud continuum. While the biological brain accomplishes complex cognition on an exceptionally modest metabolic power budget of approximately 20\,W, contemporary LLMs are profoundly energy- and memory-intensive, making sustainable lifecycle orchestration a critical operational priority.  
However, existing AI lifecycle metrics evaluate only isolated, single-model inferences or overlook multi-agent execution graphs entirely. Consequently, network operators lack foundational models to determine whether distributed agent communication incurs meaningful energy costs and where across edge-cloud tiers agent teams should physically reside. 
To address this gap, we introduce \agecal{}, generalizing the Energy Cost of AI Lifecycle (\ecal{}) metric to directed multi-agent workflows by coupling a closed-form two-rate single-call energy model (compute-bound prefill and memory-bound decode) with 7-layer OSI data transport. Grounded in hundreds of GPU benchmark configurations on NVIDIA A100 and H100, 16 open-weight models and 8 orchestration topologies, we validate components of the metric and study workflow placement implications. Our findings demonstrate that inter-agent text transport incurs $<0.25\%$ of workflow energy across 5G RAN, metro, and optical links. Therefore in edge-cloud agent  placement the dominant energy cost of distribution is often not the transmission of inter-agent text itself, but the additional inference and context processing induced by that communication. Furthermore, on an ETSI ZSM-aligned telco edge infrastructure benchmark, the evaluated multi-agent workflows increase energy by up to $23.9\times$ for Qwen2.5-7B and $4.0\times$ for Qwen3.5-9B, without producing a consistent improvement in incident remediation success.

\end{abstract}


\section{Introduction}

The biological brain accomplishes multi-step reasoning, contextual memory retrieval, and coordinated action on an exceptionally modest metabolic power budget of approximately 20\,W~\cite{mehonic2022brain}. In stark contrast, contemporary artificial intelligence (AI) based on large language models (LLMs) is profoundly energy- and memory-inefficient: state-of-the-art serving nodes draw hundreds of watts per accelerator, while per-token generation and key-value (KV) context retention rapidly saturate memory bandwidth and capacity~\cite{crawford2024generative,oviedo2026joule}. Despite these physical costs, agentic AI, where LLMs reason across multi-step loops, call external tools, retrieve domain knowledge, and coordinate as distributed teams, is increasingly embedded across the edge-cloud continuum of telecommunication networks. Because operators increasingly deploy \emph{open-weight} models distributed between user devices, edge sites, and the core, establishing a rigorous energy-memory characterization of these workloads is both fundamental and urgent.
\begin{figure}[t]
  \centering
  \includegraphics[width=\columnwidth]{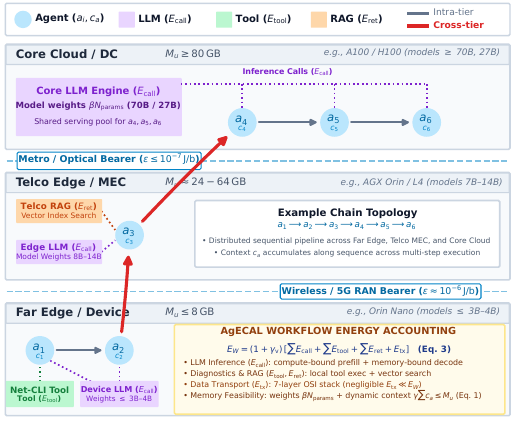}
  \caption{Agentic eCAL framework and example multi-agent workflow placement across the edge-cloud continuum.}
  \label{fig:overview}
\end{figure}

As conceptualized in Fig.~\ref{fig:overview}, an agentic workflow $W = (\mathcal{A}, \mathcal{E})$ coordinates heterogeneous operations (LLM inference with $E_{\mathrm{call}}$ energy, diagnostic tool invocations with $E_{\mathrm{tool}}$ energy, and vector database retrievals $E_{\mathrm{ret}}$ energy) across candidate network tiers. In classical distributed edge computing, placement solves an offloading trade-off between local computation and transmission energy ($E_{\mathrm{tx}}$) across wireless (5G RAN) and optical bearers. Yet, as we will show in this paper and anticipated in Fig.~\ref{fig:overview}, agentic AI fundamentally overturns this premise: inter-agent transmission is insignificant, accounts for less than $0.25\%$ of workflow energy ($E_{\mathrm{tx}} / E_{\mathrm{prefill}} \approx 1/496$ for a 2,700-token hand-off over 5G). Consequently, model size ($\beta N_{\mathrm{params}}$) and dynamic resident context ($\gamma c_a$) can dominate the energy implications of edge-cloud placement.

A placement problem is conventionally treated as an offloading trade between compute and communication energy. Yet the energy cost of such AI workflows is poorly understood: existing metrics either stop at a single
model inference or ignore the compute entirely. The original \ecal{}
metric~\cite{chou2026ecal} captured
the end-to-end energy of a \emph{single} AI model's lifecycle (i.e. data collection, preprocessing,
training, evaluation, and inference), as a per-bit quantity $[\mathrm{J/b}]$, and showed that the
more a trained model is used, the more energy-efficient each inference becomes as the fixed
development cost is amortised over a growing number of inferences. However, this no
longer describes how a growing class of network AI services are built. Instead, \emph{agentic}
workflows take a large, \emph{pre-trained, open-weight} LLM and orchestrate it typically as directed graph of: model reasoning 
over multiple steps, external tools calls, document retrieval, and cooperation with other agents as per Fig.~\ref{fig:overview}. Because such workflows are natural candidates for distribution (an agent
close to the user for latency and privacy, an agent in the core for large data corpus access), they inherit
the placement question above, and this paper asks whether the classical answer still holds:
\emph{does
distributing the agents of an LLM workflow across the network cost meaningful energy?} Three
properties make the energy accounting fundamentally different from a single
inference: 
\paragraph{No bespoke training} The model is taken off the shelf; the analogue of \ecal{}'s \cite{chou2026ecal} development energy is the model's \emph{embodied} (pre-training) energy, which is the most significant but shared across all of the model's users.
\paragraph{The unit of work is a workflow, not an inference} A single task triggers a graph of
  many LLM calls, whose prompts \emph{carry the accumulating transcript}, so token volume, and
  hence energy, grows super-linearly in the number of steps.

\paragraph{Heterogeneous components} Beyond LLM inference, agentic execution incorporates external tool execution ($E_{\mathrm{tool}}$), vector database retrieval ($E_{\mathrm{ret}}$), and 7-layer OSI transmission ($E_{\mathrm{tx}}$), as depicted in Fig.~\ref{fig:overview}.

Our contributions are:
\begin{itemize}
\item We introduce \agecal{}, a rigorous per-bit energy framework for multi-agent workflows executing across the edge-cloud continuum, summing LLM inference, tool execution, vector retrieval, 7-layer OSI transmission, and amortized embodied costs (Sec.~\ref{sec:metric}). \agecal{} generalizes \ecal{}~\cite{chou2026ecal}, reducing to its inference metric for one-shot calls.
  
\item We establish a physics- and hardware-grounded two-rate closed-form energy model for individual LLM calls (Secs.~\ref{sec:call} and ~\ref{sec:callseq}), decomposing inference into compute-bound prefill and memory-bandwidth-bound decode regimes, parameterized by serving batch amortization and context-dependent KV-cache traffic.
  
\item We validate the framework against extensive GPU measurements on NVIDIA A100 and H100 with vLLM continuous batching ($R^2 \ge 0.99$, MAPE $\approx 10\%$). Systemic characterization reveals that: (i) history-carrying loops exhibit super-linear prompt-prefill energy scaling; (ii) inter-agent transmission across 5G RAN, metro, and optical links accounts for $<0.25\%$ of workflow energy; and (iii) on an ETSI ZSM-aligned telco edge infrastructure benchmark, compared with the single-agent baseline, the evaluated multi-agent topologies increase energy consumption by up to $23.9\times$ for Qwen2.5-7B and $4.0\times$ for Qwen3.5-9B, without producing a consistent improvement in incident remediation success rates\footnote{The code will be open sourced upon acceptance. https://github.com/sensorlab/a-eCAL}.
\end{itemize}

\textit{Paper Organization.} The remainder of this paper is structured as follows. Sec.~\ref{sec:related} surveys related work across AI lifecycle metrics, LLM serving energy, and multi-agent systems. Sec.~\ref{sec:metric} formalizes the \agecal{} metric and system model, deriving the closed-form two-rate single-call energy model,  super-linear context accumulation, and the 7-layer OSI transmission formulation. Sec.~\ref{sec:results-scale} presents empirical GPU validation and scaling laws across reasoning depth, agent count, and batching. Sec.~\ref{sec:results-comms} evaluates multi-agent placement feasibility, accelerator memory boundaries, and transmission energy across network tiers. Sec.~\ref{sec:infra} evaluates the framework on an ETSI ZSM-aligned telco edge infrastructure benchmark. Sec.~\ref{sec:discussion} discusses limitations and future extensions, and Sec.~\ref{sec:conclusion} concludes the paper.

\section{Related Work}\label{sec:related}
\textit{Lifecycle energy metrics.} \ecal{}~\cite{chou2026ecal} introduced a per-bit,
component-resolved view of AI energy spanning the OSI and cloud-computing reference architectures.
We adopt the same per-bit framing and the same transmission/virtualization machinery, and extend the
\emph{development} and \emph{inference} components to agentic LLM workflows.

\textit{LLM inference compute and energy.} The compute of a decoder-only transformer is governed by
a well-established FLOP accounting: a forward (inference) pass costs $\approx 2N_{\mathrm{params}}$ FLOPs per token and
a full training step $\approx 6N_{\mathrm{params}}$ (the backward pass being roughly twice the forward), with a
context-dependent attention term that is negligible until the context length exceeds
$\approx 8d$~\cite{kaplan2020scaling,deepmind_scaling_book}. On the measurement side, Samsi
\emph{et al.}~\cite{samsi2023words} report $3$--$4$\,J per output token for LLaMA~65B on V100/A100,
the ML.ENERGY benchmark~\cite{mlenergy2025} shows per-generation energy falling $\sim\!3.5\times$
from batch~4 to batch~64 on H100 and warns that estimating energy from nameplate TDP overestimates
measured energy by up to $4.1\times$, and frontier-scale serving reaches a median $0.31$\,Wh per
query~\cite{oviedo2026joule}. We adopt the $\beta
 N_{\mathrm{params}}$ with $\beta=2$ accounting and calibrate a utilization factor to
these measurements rather than to peak FLOP/s.

\textit{Embodied cost of open-weight models.} Pre-training energy is documented in primary model
cards and peer-reviewed audits: Llama~2 required $3.3$M A100-hours ($539$\,tCO$_2$eq) and Llama~3
$7.7$M H100-hours ($2290$\,tCO$_2$eq, of which $1.3$M\,h for the 8B and $6.4$M\,h for the
70B)~\cite{touvron2023llama2,llama3card}; Luccioni \emph{et al.}~\cite{luccioni2023bloom} decompose
BLOOM-176B's $50.5$\,tCO$_2$eq lifecycle into $\sim\!22\%$ embodied and $\sim\!78\%$ operational.
Crucially, inference only overtakes training at fleet scale, with parity after $\sim\!205$M
(BLOOMz-560M) to $\sim\!593$M (BLOOMz-7B) inferences~\cite{luccioni2024power}, which motivates
amortizing embodied
energy over deployment volume.

\textit{Agentic, reasoning, and RAG systems.} The defining energy property of agentic AI is
\emph{token amplification}: reasoning models emit \reasoningBlow{} more tokens than standard models on
the same task (e.g.\ Phi-4-reasoning-plus $6{,}780$ vs Phi-4 $378$ tokens), sometimes at lower
accuracy~\cite{llmthinkbench2025}, and test-time scaling with $\sim\!15\times$ tokens raises per-query
energy $\sim\!13\times$~\cite{oviedo2026joule}. Because energy scales near-linearly with generated
tokens, this directly multiplies the per-token term. Agent composition compounds it: single agents
use $\sim\!4\times$ and multi-agent systems \multiagentx{} more tokens than a chat
call~\cite{anthropic_multiagent}, and multi-agent debate can reach $32$--$158\times$ a single call
before sparse-communication methods recover much of it~\cite{s2mad2025}. In retrieval-augmented
generation the injected context dominates: a $\sim\!100$-token request plus $\sim\!1000$ retrieved
tokens makes the augmented request $>\!10\times$ costlier~\cite{ragcache2024}, while prefix/KV-cache
reuse of that context cuts time-to-first-token up to $4\times$~\cite{ragcache2024} and tuned RAG
pipelines save up to $60\%$~\cite{ragenergy2026}.
Model choice is a complementary lever: measured per-token energy rises
$\sim\!18\times$ from Llama-3 1B to 70B, so a 1B model costs roughly $5\%$
of a 70B model per generated token~\cite{tokenpowerbench2026},
as is sparsity: Mixture-of-Experts
energy tracks \emph{active} not total parameters, e.g.\ a 30B-A3B model uses $3.56\times$ less
energy/token than a dense 32B~\cite{mlenergy_v3}. Crucially, all agentic/tool evidence is in
\emph{token} units, and no published study measures the joules of a full open-weight agentic run,
so we bridge tokens to energy via the per-token model and flag the tool/orchestration terms as
analytical.

\textit{Agentic AI in network management.} Modern agentic AI has the potential to deliver the autonomy that the zero-touch literature specifies: ETSI ZSM defines closed management loops over network
domains~\cite{etsi_zsm002}, and O-RAN places those loops on a virtualized cloud substrate spanning
far-edge, edge and regional sites~\cite{oran_wg6_cloud}. As LLM-based agents are
adopted as the reasoning element of such loops, their energy becomes an operational quantity rather
than an academic one. Energy accounting for the AI/ML workflow inside O-RAN has so far addressed
\emph{single-model} pipelines~\cite{chou2025oran,chou2024standardization}; we extend it to the
multi-call, multi-agent case.

\textit{Placement and orchestration.} Distributing computation across network tiers is
classically an offloading trade, in which compute energy saved by moving work outward is repaid in
transmission\cite{mach2017mobile}.
Our measurements indicate that this trade does not arise for text hand-offs between agents: what a
hand-off carries is small in bits but large in the prefill it triggers on arrival, an asymmetry
consistent with the load-independence of router power~\cite{jacob2025routers}. Orchestration architectures for teams of agents have been analyzed from the perspective of collaboration modes, roles, models, or workflows per query \cite{yue2025masrouter,zhang2025maas,yu2026adaptorch}, and communication-efficient multi-agent reasoning has been pursued by sparsifying which agents address
which~\cite{s2mad2025}; we evaluate the graph instead for the traffic its edge count implies and
for the memory its retained context occupies.

\begin{table}[t]
\caption{Summary of mathematical notation and parameters.}
\label{tab:notation}
\centering
\small
\setlength{\tabcolsep}{3.5pt}
\renewcommand{\arraystretch}{1.18}
\begin{tabular}{@{}l >{\raggedright\arraybackslash}p{6.6cm}@{}}
\toprule
Symbol & Description \\
\midrule
$W=(\mathcal{A},\mathcal{E})$ & workflow DAG: agents $\mathcal{A}$, directed edges $\mathcal{E}$ ($E=|\mathcal{E}|$) \\
$N,\, K$ & team width ($N=|\mathcal{A}|$), workflow execution depth (steps) \\
$x_{au} \in \{0,1\}$ & placement indicator: agent $a$ at site $u \in \mathcal{D}$ \\
$M_u,\, b_u$ & accelerator VRAM capacity [B], serving batch at site $u$ \\
$c_a,\,\gamma$ & context length of agent $a$ [tokens], KV-cache bytes/token \\
$\beta,\, N_{\mathrm{params}}$ & bytes/param ($\beta{=}2$), total parameters ($N_{\mathrm{act}}$: MoE) \\
$n_\ell,\, d$ & number of transformer layers, hidden dimension size \\
$p_{\mathrm{in}},\,p_{\mathrm{out}}$ & prompt (prefill) and generated (decode) token counts \\
$C_{\mathrm{call}},\,\Pi$ & forward pass FLOPs, peak throughput [FLOP/s] \\
$\eta,\, P$ & model-FLOP utilization factor, board operational power [W] \\
$c_{\mathrm{pre}},\,c_{\mathrm{dec}}(b)$ & per-token prefill and decode energy rates [J/token] \\
$a_{\mathrm{dec}},\,c_0$ & decode streaming slope, asymptotic KV floor [J/token] \\
$E_W,\, E_{\mathrm{call}}$ & operational workflow energy, single LLM call energy [J] \\
$E_{\mathrm{tool}},\,E_{\mathrm{ret}}$ & execution energy of tool calls and vector retrieval [J] \\
$E_{\mathrm{tx}},\,\varepsilon_{uv}$ & inter-agent transmission energy [J], bearer intensity [J/b] \\
$B_{\mathrm{payload}},\,B_{\mathrm{T},p}$ & application payload [b], physical transmitted bits [b] \\
$OH_{\mathrm{DP}},\,RR_{\mathrm{DP}}$ & protocol overhead ratio, wireless retransmission ratio \\
$E_{\mathrm{emb}},\,G$ & pre-training embodied energy [J], lifetime invocations \\
$\gamma_{\mathrm{v}},\,\gamma_{\mathrm{e}}$ & virtualization factor, embodied amortization fraction ($1/G$) \\
$B_{\mathrm{useful}},\,f$ & useful workflow output bits ($f \cdot T_{\mathrm{out}}$), bits per token \\
\bottomrule
\end{tabular}
\end{table}

\section{Agentic Workflows and the Agentic-eCAL Metric}\label{sec:metric}
We define an agentic workflow as a directed graph
$W=(\mathcal{A},\mathcal{E})$ of $N$ agents joined by $E=|\mathcal{E}|$ hand-offs and executed in
$K$ steps; step $k$ is run by agent $a(k)\in\mathcal{A}$ and is one LLM call of
$p_{\mathrm{in}}^{(k)}$ prompt and $p_{\mathrm{out}}^{(k)}$ generated tokens, optionally with a
retrieval before it and a tool invocation after. As per the conceptual illustration in Fig. \ref{fig:overview}, it is hosted on a set $\mathcal{D}$ of candidate
deployment sites spanning various network segments from the user device to the core: site $u$ offers memory $M_{u}$ for
weights and cache and sustains a serving batch $b_{u}$, and sites $u,v$ are joined by a bearer of
energy intensity $\varepsilon_{uv}$ [J/b], with
$\varepsilon_{uu}=0$ for co-located agents. A deployment of a workflow is then the triple
$\mathcal{C}=(x,\pi,\mathcal{E})$: the \emph{placement} $x_{au}\in\{0,1\}$, one if agent $a$ runs at
site $u$, with $\sum_{u}x_{au}=1$; the \emph{hand-off protocol} $\pi\in\mathcal{P}$, which fixes the
payload $\theta_{\pi}(a,a')$ in bits each edge carries, whether the accumulated transcript, an
increment, a bounded summary or the key-value cache itself; and the \emph{architecture}
$\mathcal{E}$, which fixes how many edges there are.

Agent role types vary across the sources this paper draws on, so we keep each source's names and
classify agents by function instead in four classes as follows. \emph{Generate} agents produce a
candidate or a plan, and are charged principally for $p_{\mathrm{out}}$. \emph{Transform} agents
refine a single line of work. \emph{Ground} agents acquire state external to $W$, through
retrieval or a tool, and so import into $p_{\mathrm{in}}$ a segment no agent generated.
\emph{Aggregate} agents reduce a set of competing candidates and read all of it, so their
$p_{\mathrm{in}}$ grows with the size of that set.

We define a placement configuration as admissible only if each site $u \in \mathcal{D}$ can hold the weights of the
model it runs together with the resident context of every agent it hosts,
\begin{equation}
\underbrace{\beta N_{\mathrm{params}}}_{\text{static model weights}} \;+\; \underbrace{\gamma \sum_{a \in \mathcal{A}} x_{au} c_a}_{\text{dynamic resident context}} \;\le\; M_u, \qquad \forall u \in \mathcal{D},
\label{eq:c-mem}
\end{equation}
where the first term is the static memory footprint of model weights loaded once per site at $\beta$ bytes per parameter, and the second term is the dynamic footprint of the key-value (KV) cache, requiring $\gamma$ bytes per context token $c_a$ for each agent $a$ hosted at site $u$ ($x_{au}=1$). While model weights load once, KV caches scale with agent concurrency and retained context length.

To quantify the energy of an agentic workflow, a new metric is required. For a workflow $W$, we generalize the  \ecal{} metric~\cite{chou2026ecal}, by  defining \agecal{} as the ratio of total energy consumed to the useful application-level data produced:
\begin{equation}
\agecal = \frac{E_W + \gamma_{\mathrm{e}} (E_{\mathrm{emb}} + E_{\mathrm{emb,ret}})}{B_{\mathrm{useful}}} \quad [\mathrm{J/b}],
\label{eq:metric}
\end{equation}
where $E_W$ is the operational energy of the workflow, $E_{\mathrm{emb}}$ is the embodied
(pre-training) energy of the open-weight model, $E_{\mathrm{emb,ret}}$ is the embodied
(pre-training) energy of the open-weight model used when the optional retrieval step is included, $\gamma_{\mathrm{e}}\in[0,1]$ is the fraction of that
embodied energy attributed to one workflow invocation, and
$B_{\mathrm{useful}} = f\,T_{\mathrm{out}}$ is the useful output measured in bits ($T_{\mathrm{out}}$
useful output tokens at $f$ bits/token). Table~\ref{tab:notation} summarizes all the notations in the paper.

\paragraph{The operational energy of a workflow $E_{W}$} sums the energy of every component, scaled by a virtualization and system 
orchestration overhead $\gamma_{\mathrm{v}}$ (the same factor \ecal{} uses for cloud platform
overheads, Eq.~(30) of~\cite{chou2026ecal}):
\begin{equation}
E_W = (1+\gamma_{\mathrm{v}})\!\left[\sum_{k=1}^{K} E_{\mathrm{call}}^{(k)} + \sum_{k} E_{\mathrm{tool}}^{(k)} + \sum_{k} E_{\mathrm{ret}}^{(k)} + E_{\mathrm{tx}}\right]
\label{eq:ew}
\end{equation}
\noindent where $E_{\mathrm{call}}^{(k)}$ is the energy of the LLM call at step k, $E_{\mathrm{tool}}^{(k)}$ the execution energy of any tool that step invokes, $E_{\mathrm{ret}}^{(k)}$ the cost of any retrieval preceding it (query embedding plus index search), and $E_{\mathrm{tx}}$ the transmission energy of the messages received from other  agents. At $\!K\!=\!1$ with no tools or retrieval and with co-located agents, $E_W\!\to\!(1{+}\gamma_{\mathrm{v}})E_{\mathrm{call}}$ and
Eq.~\ref{eq:metric} reduces to \ecal{}'s energy-per-bit of a single inference, with the denominator
specialising \ecal{}'s \emph{manipulated} application level data bits to the workflow's \emph{useful output} bits. \agecal{} is thus a strict generalization of
\ecal{}, adding the multi-call graph on top of the single-inference term.

\paragraph{The embodied energy of workflow} If a model serves $G$ agentic workflow invocations over its deployed lifetime, the
embodied energy attributable to one invocation is $\gamma_{\mathrm{e}} (E_{\mathrm{emb}} + E_{\mathrm{emb,ret}})$ from Eq. \ref{eq:metric} where  $\gamma_{\mathrm{e}}=1/G$ refers to the fraction of the energy from training the LLM models used for answering the calls $E_{\mathrm{emb}}$ and for retrieval when this is present $E_{\mathrm{emb,ret}}$. The more workflows an LLM powers, the lower the fixed training energy attributable to that workflow. 

The existing \ecal{} \cite{chou2026ecal} amortizes a model's one-time development energy $E_{\mathrm{D}}$ over the number of
inferences served. For an open-weight model the development cost is the \emph{pre-training} energy
$E_{\mathrm{emb}}$, which is incurred once by the model producer but shared across \emph{all}
downstream deployments. As with \ecal{}'s amortization-over-inferences result, this share
vanishes at scale, so for widely served open-weight models the operational term $E_W$ dominates
\agecal{}. The crossover point (how many invocations are needed before operation overtakes embodied cost) depends mostly on $E_{\mathrm{emb}}$.

\subsection{Energy of a Single LLM Call: $E_{\mathrm{call}}$}\label{sec:call}
An LLM call executes in two distinct phases, yielding a \emph{two-rate} energy model:
\begin{equation}
E_{\mathrm{call}}(p_{\mathrm{in}},p_{\mathrm{out}};b)\;\approx\;c_{\mathrm{pre}}\,p_{\mathrm{in}}\;+\;c_{\mathrm{dec}}(b)\,p_{\mathrm{out}},
\label{eq:tworate}
\end{equation}
where $c_{\mathrm{pre}}$ is the per-prefill-token energy and $c_{\mathrm{dec}}(b)$ is the per-decode-token energy at serving batch size $b$. 

\textit{Prefill} processes all $p_{\mathrm{in}}$ prompt tokens in parallel. It is compute-bound at near-peak utilization ($\eta_{\mathrm{pre}}$) and, being a single batched matrix multiplication, is essentially independent of $b$:
\begin{equation}
c_{\mathrm{pre}}\approx N_{\mathrm{params}}\beta\,P/(\eta_{\mathrm{pre}}\Pi)
\label{eq:pre}
\end{equation}
\textit{Decode} emits $p_{\mathrm{out}}$ tokens sequentially. It is memory bandwidth bound accounting for $77$--$91\%$ of inference time~\cite{decode_membound2025} as weights and the growing KV-cache stream from High Bandwidth Memory (HBM) while arithmetic units sit largely idle. Because weights are read once per batch, this cost amortizes over the $b$ concurrently-decoding sequences:
\begin{equation}
c_{\mathrm{dec}}(b)\;\approx\;\frac{P}{B_{\mathrm{HBM}}}\!\left(\underbrace{\frac{N_{\mathrm{params}}\beta}{b}}_{\text{weights}/\text{batch}}
   \;+\;\underbrace{\gamma\,\bar c}_{\text{KV-cache}}\right),
\label{eq:cdec}
\end{equation}
\noindent where $B_{\mathrm{HBM}}$ is HBM bandwidth,\footnote{In practice a serving stack sustains
only about $70\%$ of the theoretical peak, so Eq.~\ref{eq:cdec} at the nameplate
$B_{\mathrm{HBM}}$ is a lower bound on $c_{\mathrm{dec}}$. Model-bandwidth utilization,
$(\text{achieved bandwidth})/(\text{peak bandwidth})$, is measured at $72\%$ for Llama-7B in fp16
on an A100-80GB, where the ceiling is itself under $85\%$ because ``even just copying memory
struggles to break'' it~\cite{pytorch2023gptfast}, and at $60\%$ and $55\%$ on $2{\times}$H100-80GB
and $4{\times}$A100-40GB~\cite{databricks2023inference}. All three are single-stream figures;
utilization falls further as $b$ grows and attention over the accumulated cache displaces the
weight read.} $\beta$ is bytes per parameter, and $\gamma\bar c$ is the
per-token KV-cache traffic at mean context $\bar c$.

This two-rate formulation reveals an important  crossover. Mathematically, a call becomes prefill-dominated when $c_{\mathrm{pre}}\,p_{\mathrm{in}} > c_{\mathrm{dec}}(b)\,p_{\mathrm{out}}$, which occurs when the token ratio $p_{\mathrm{in}}/p_{\mathrm{out}} > c_{\mathrm{dec}}(b)/c_{\mathrm{pre}}$. At small batches, weight-loading dominates ($c_{\mathrm{dec}} \gg c_{\mathrm{pre}}$) and energy tracks the linear decode workload. However, as $b$ scales, the weight term in Eq.~\ref{eq:cdec} vanishes and $c_{\mathrm{dec}}(b)$ decreases  toward the KV-cache floor, rapidly driving down the right side of the inequality. This dynamic fundamentally shapes agentic AI: because transcripts accumulate across steps, production-batched workflows inevitably cross this threshold into the prefill-dominated regime. Consequently, the workflow's energy scales super-linearly, whereas the exact same workflow under single-stream serving would remain linear.

\subsection{Energy of the Workflow LLM Calls: $E_{\mathrm{call,w}}$}\label{sec:callseq}
In a history-carrying loop the prompt of step $k$
carries the system/tool-schema prompt $p_{\mathrm{sys}}$, any newly introduced tokens, retrieved
context, and the accumulated transcript $h_k=\sum_{j<k}\!\big(p_{\mathrm{out}}^{(j)}+o^{(j)}\big)$
($o^{(j)}$: tool observations). With a fixed emission $s$ per step, $p_{\mathrm{in}}^{(k)}\!\approx\!
p_{\mathrm{sys}}+s\,k$, so the \emph{token} volume is exactly quadratic while the generated output is
linear:
\begin{equation}
\sum_{k=1}^{K} p_{\mathrm{in}}^{(k)} = \tfrac{s}{2}K^2 + O(K),
\qquad
\sum_{k=1}^{K} p_{\mathrm{out}}^{(k)} = sK .
\label{eq:tokgrow}
\end{equation}
Substituting into Eq.~\ref{eq:tworate} through the two-rate model, the operational energy is
\begin{equation}
E_{call,w} \;\propto\; c_{\mathrm{pre}}\,\big(\tfrac{s}{2}K^2\big) \;+\; c_{\mathrm{dec}}(b)\,sK ,
\label{eq:callscale}
\end{equation}
so \agecal{} scales as $K^{a}$ with a \emph{batch-set} exponent $1\!\le\!a\!\le\!2$: at small batch
$c_{\mathrm{dec}}(b)\!\gg\!c_{\mathrm{pre}}$ and the linear decode term dominates ($a\!\to\!1$); as $b$
grows, $c_{\mathrm{dec}}(b)\!\to\!0$ and the quadratic prefill term takes over ($a\!\to\!2$). Eq.~\ref{eq:tokgrow} shows that the \emph{context} grows quadratically unconditionally, but as Eq. (\ref{eq:callscale}) shows, the \emph{energy of a sequence of calls} inherits that growth only when prefill-dominated. This quadratic scaling governs stateless agent
invocations across distributed network hosts or microservices
without shared memory, where each step must re-prefill the
accumulated prompt history. While stateful single-host serving engines leverage Automatic Prefix Caching (APC) to reduce redundant prompt prefill to linear scaling O(K) for co-located calls, distributed edge-cloud agent pipelines that dispatch steps
across distinct edge servers or microservices may incur full transcript re-prefill penalties.

\subsection{Energy of Tool Calls: $E_{\mathrm{tool}}$} 
A tool invocation contributes its own execution energy $E_{\mathrm{tool}}^{(k)}$
and injects $o^{(k)}$ observation tokens into the transcript. Non-LLM tools that fetch data over the
network are priced by the OSI data-collection model $E_{DC}$ of \ecal{}~\cite{chou2026ecal}); local compute tools by their FLOPs and the same energy mapping as
Eq.~\ref{eq:pre}.

\begin{figure*}[!t]
  \centering
  \includegraphics[width=0.8\textwidth]{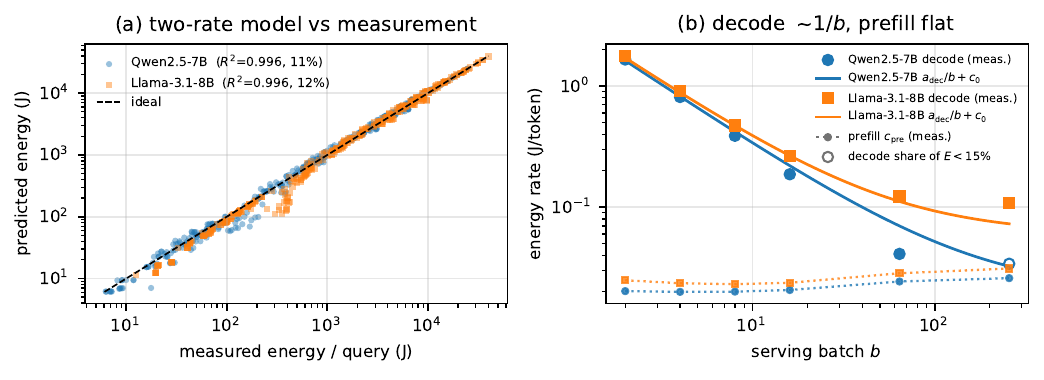}
  \caption{Empirical validation of the two-rate model (Qwen2.5-7B and Llama-3.1-8B; A100, vLLM;
  $270$ configurations per model, $2$--$30$ agents, serving batch $b\!\in\![2,256]$).
  (a)~Predicted vs.\ measured per-query GPU energy  ($R^2\!>\!0.99$, mean absolute error ${\sim}10\%$); (b)~The two rates
  recovered by regressing energy on $(p_{\mathrm{in}},p_{\mathrm{out}})$: the prefill rate $c_{\mathrm{pre}}$ is batch-independent, while the decode rate collapses as $a_{\mathrm{dec}}/b$ toward a small floor (Eq.~\ref{eq:cdec}). Llama's higher floor
  keeps it partly decode-bound at large batch. }
  \label{fig:validation}
\end{figure*}

\subsection{Energy of Retrieval: $E_{\mathrm{ret}}$} 
A retrieval step embeds the query with an embedding model
($\approx \beta N_{\mathrm{emb}}\,p_q$ FLOPs for $p_q$ query tokens), searches a vector index of
$M_{\mathrm{vec}}$ vectors of dimension $d_{\mathrm{emb}}$ (cost $\propto d_{\mathrm{emb}}M_{\mathrm{vec}}$
for exact search, or $\propto d_{\mathrm{emb}}\log M_{\mathrm{vec}}$ for graph-based approximate
search)~\cite{HNSW_Malkov2020}, and injects $k_r$ chunks of $L_r$ tokens that enlarge the next
call's prompt. By mapping the embedding energy via Eq.~\ref{eq:pre} and index search operations via memory bandwidth yields:
\begin{equation}
E_{\mathrm{ret}}^{(k)} = \frac{\beta N_{\mathrm{emb}}\,p_q\,P_{\mathrm{emb}}}{\eta_{\mathrm{emb}}\,\Pi_{\mathrm{emb}}} \;+\; \alpha_{\mathrm{idx}}\,d_{\mathrm{emb}}\log(M_{\mathrm{vec}}),
\label{eq:eret}
\end{equation}
\noindent where $\alpha_{\mathrm{idx}}$ characterizes index distance calculation energy [J/op]. The added context tokens are
captured by
Eqs.~\ref{eq:tokgrow}. Empirically, growing the context from $2$K to $10$K tokens raises measured energy per token
$\sim\!3\times$ for a $70$B model~\cite{tokenpowerbench2026,sweetspot2026}. 

\subsection{Energy of Transmission for Multi-agent Communication: $E_{\mathrm{tx}}$}\label{sec:comm}
When agents are co-located on the same physical host, the hand-off cost is strictly confined to local memory transfers and additional prompt tokens. When agents are distributed across network tiers (e.g., edge devices, base station MEC servers, metro aggregation sites, or central cloud datacenters), inter-agent messages incur data communication energy. In Eq.~\ref{eq:ew}, the total transmission energy is $E_{\mathrm{tx}} = \sum_{(u,v)\in\mathcal{E}} B_{\mathrm{payload}, uv} \cdot \varepsilon_{uv}$, where $B_{\mathrm{payload}, uv}$ is the application payload volume in bits along directed edge $(u,v)$, and $\varepsilon_{uv}$ is the effective end-to-end transport energy intensity [J/bit].

Following the model in the eCAL~\cite{chou2026ecal} framework, the actual bits transmitted over the network for $B_{\mathrm{payload}, uv}$ can be computed with:
\begin{align}
B_{\mathrm{T}, p}&~[b] =\lceil B_{\mathrm{T}, L_{\mathrm{OSI}}, p} \prod_{l=1}^{L_{\mathrm{OSI}}} \underbrace{\left(RR_{\mathrm{DP},l}  (1 + OH_{\mathrm{DP},l})\right.}_{\text{contribution from DP} }\nonumber\\
&+ \underbrace{\left.RR_{\mathrm{CP},l} \gamma_l   OH_{\mathrm{CP},l}\right)}_{\text{contribution from CP}} \rceil,~\mathrm{where}~L_{\mathrm{OSI}}=7,
\label{eq:bit_expansion}
\end{align}
where $OH_{\mathrm{DP}, l}$ is the protocol overhead ratio at layer $l$ (encompassing L2 MAC framing, L3 IP routing, L4 TCP headers, L6 TLS 1.3 encryption, and L7 HTTP/2 or gRPC protobuf serialization, typically adding $100\%$ to $200\%$ overhead), and $RR_{\mathrm{DP}, l}$ is the retransmission ratio. For wireless cellular links (e.g., 5G NR), channel fading and interference induce Hybrid Automatic Repeat reQuest (HARQ) retransmissions, yielding $RR_{\mathrm{DP}} \approx 1.5$ to $2.0$ under cell-edge or high-mobility channel conditions.

The same framework establishes the total energy $E_{\mathrm{DC}, p}$ required to transport payload $B_{\mathrm{payload}}$ across physical link $p$ sums RF transmission/reception with protocol execution:
\begin{align}
E_{\mathrm{DC}, p} = &\underbrace{\frac{P_{\mathrm{T}, p}}{R_{\mathrm{T}, p}} B_{\mathrm{T}, p}}_{\text{PHY transmit } E_{\mathrm{T}, p}} + \underbrace{\frac{P_{\mathrm{R}, p}}{R_{\mathrm{R}, p}} B_{\mathrm{T}, p}}_{\text{PHY receive } E_{\mathrm{R}, p}} \nonumber\\
&+ \sum_{l=2}^{L_{\mathrm{OSI}}} \Big( \underbrace{B_{\mathrm{T}, l, p} \cdot N_{\mathrm{dev}, l, p} \cdot P_{\mathrm{dev}, p}}_{\text{L2 to L7 protocol processing at end device}} \nonumber\\
&\qquad\quad + \underbrace{B_{\mathrm{T}, l, p} \cdot N_{\mathrm{gw}, l, p} \cdot P_{\mathrm{gw}, p}}_{\text{L2 to L7 protocol processing at gateway}} \Big),
\label{eq:E_DC_final}
\end{align}
where $P_{\mathrm{T}, p}$ and $P_{\mathrm{R}, p}$ are the physical transceiver powers [W], $R_{\mathrm{T}, p}$ and $R_{\mathrm{R}, p}$ the operational data rates [b/s], $N_{\mathrm{dev}, l, p}$ and $N_{\mathrm{gw}, l, p}$ the CPU clock cycles per bit for protocol processing at device and gateway, and $P_{\mathrm{dev}, p}, P_{\mathrm{gw}, p}$ the computational energy efficiencies [J/cycle]. 

%

\begin{table}[t]
\caption{Model vs.\ measured per-output-token energy, each row evaluated at the serving
batch its source reports.}
\label{tab:validation}
\centering
\small
\begin{tabular}{@{}lllll@{}}
\toprule
Model & HW & $b$ & $E_{\mathrm{call}}$ & measured \\
\midrule
%
Llama-3 70B & H100 NVL & $128$ & \jpertokSeventyBthru{} & \jpertokSeventyBmeas{}~\cite{tokenpowerbench2026} \\
Llama-3 8B & H100 NVL & $128$ & \jpertokEightBthru{} & \jpertokEightBmeas{}~\cite{tokenpowerbench2026} \\
Qwen2.5-7B  & H100 NVL & $128$ & \jpertokQwenthru{} & \jpertokQwenmeas{}~\cite{tokenpowerbench2026} \\
\bottomrule
\multicolumn{5}{l}{\footnotesize Units: J per output token; model and measurement both GPU-only.} \\
\multicolumn{5}{l}{\footnotesize Measured: GPU-only, $0$--$2$K context band, of~\cite{tokenpowerbench2026} ($4{\times}$H100 node).} \\
\multicolumn{5}{l}{\footnotesize The 7B and 8B runs occupy one GPU, so ${\sim}38\%$ of their} \\
\multicolumn{5}{l}{\footnotesize  value is idle sibling draw ($\approx\!0.07$); the 70B is sharded over all four.} \\
\multicolumn{5}{l}{\footnotesize   Their published $0.11$ is therefore quoted here at the active device,} \\
\multicolumn{5}{l}{\footnotesize  $0.11{\times}0.62$, the share of node GPU power that device draws} \\
\multicolumn{5}{l}{\footnotesize  across the $74$ released runs; the $70$B needs no such correction.}
\end{tabular}
\end{table}


\section{Energy and Memory Characterization of Agentic Workflows }\label{sec:results-scale}

This section aims at experimentally and numerically validating \agecal{} component by component, as introduced in Sec. \ref{sec:metric}. We first calibrate and test the single-call
model of Eq.~\ref{eq:tworate} against direct GPU energy measurement, then examine the scaling law
it implies for a sequence of calls Eq. \ref{eq:callscale}, then apply the composed metric to a complete workflow Eq. \ref{eq:ew}, and
finally consider how the embodied term amortizes Eq. \ref{eq:metric}.

\subsection{Experimental Validation of the LLM Call Energy Model} \label{sec:singlecall}

Eq. \ref{eq:ew} breaks down the energy of a workflow $E_W$ per component, while Sec. \ref{sec:call} focuses on the first term $E_{call}$ and introduces a two rate energy model of a single LLM call in Eq. \ref{eq:tworate}. To validate the model in Eq. \ref{eq:tworate}, we vary the number of agents $N$, steps $K$ and batch size $b$. We select two open weight 7-8bn LLMs, namely Qwen2.5-7B, and Llama-3.1-8B (NVIDIA A100, vLLM continuous
batching), across team sizes of $2$ to $30$ agents. Energy is metered with NVML. As depicted in Fig.~\ref{fig:validation}a, the energy provided by Eq. \ref{eq:tworate} (y axis) closely matches the measurements (x axis) with 
$R^2\!>\!0.99$ and mean absolute error ${\sim}10\%$. 

We notice that  the two terms of Eq.~\ref{eq:cdec} can also be expressed as 
$c_{\mathrm{dec}}(b)=a_{\mathrm{dec}}/b+c_{0}$, in which $a_{\mathrm{dec}}$ is the weight-read
coefficient, the part that amortises over the batch, and $c_{0}$ the KV-cache floor, which does
not. By regressing the measured energies at each batch we find  $a_{\mathrm{dec}}=3.2$ and $3.3$
and $c_{0}=0.02$ and $0.06$\,J/token for Qwen2.5-7B and Llama-3.1-8B respectively. 
The fitted
$a_{\mathrm{dec}}$ lies a factor of about $2.7$ below the datasheet value of the terms of 
Eq.~\ref{eq:cdec}. We therefore read that equation
as establishing the form of the two terms rather than as
predicting their magnitude to better than a third. We plot the curves with these coefficients together with the measurements in Fig.~\ref{fig:validation}b and used throughout.

It can be seen from Fig.~\ref{fig:validation}b that the measurements validate the two rate model from Eqs. \ref{eq:pre} and \ref{eq:cdec}: a near-constant $c_{\mathrm{pre}}\!\approx\!0.02$--$0.03$\,J/token
(the compute-bound prefill floor) and a decode rate that decreases with  $\sim\!1/b$ ($p\!<\!0.001$). Both
rates match first principles: $c_{\mathrm{pre}}$ implies a prefill utilization
$\eta_{\mathrm{pre}}\!\approx\!0.74$--$0.85$
(compute-bound, as expected), and the decode coefficient agrees with the memory-bandwidth prediction
$PN\beta/B_{\mathrm{HBM}}$ to within $10\%$. 

As an independent
cross-hardware check, Table~\ref{tab:validation} evaluates Eq.~\ref{eq:tworate} on hardware and
at serving batches we did not measure ourselves, from datasheet quantities alone. Considering input from the 
benchmark, including its Standard Load profile ($b\!=\!128$, $p_{\mathrm{out}}\!=\!500$), its Alpaca
prompts and its $4{\times}$H100 NVL node, the model returns \jpertokSeventyBthru{}\,J/token for
Llama-3~70B, \jpertokEightBthru{} for Llama-3~8B and \jpertokQwenthru{} for Qwen2.5-7B, against
measured \jpertokSeventyBmeas{}, \jpertokEightBmeas{} and
\jpertokQwenmeas{}~\cite{tokenpowerbench2026}. The two smaller models occupy one device of four,
so their published figures are corrected to that device before comparison; while the $70$B shards over
all four and needs no correction. The check shows that the model underestimates $E_{call}$ of the smaller models and overestimates for the larger one. This discrepancy is  consistent across a tenfold range of model size and lies
in the magnitude of Eq.~\ref{eq:cdec} and not its form: evaluated at the
datasheet $B_{\mathrm{HBM}}$ the decode rate is a lower bound on served energy,
because a serving stack sustains only ${\sim}70\%$ of peak bandwidth even in the
favourable single-stream case (footnote to Eq.~\ref{eq:cdec}) and the equation
omits the empirical decode floor $c_0$, KV-cache fragmentation, idle
draw and the cost of attention over the accumulating cache. This shortfall
widens with batch, which is why the coefficients calibrated at our serving
batches (Fig.~\ref{fig:validation}) sit above the datasheet curve and are the
ones we use throughout.
\begin{figure*}[t]
  \centering
  \includegraphics[width=\textwidth]{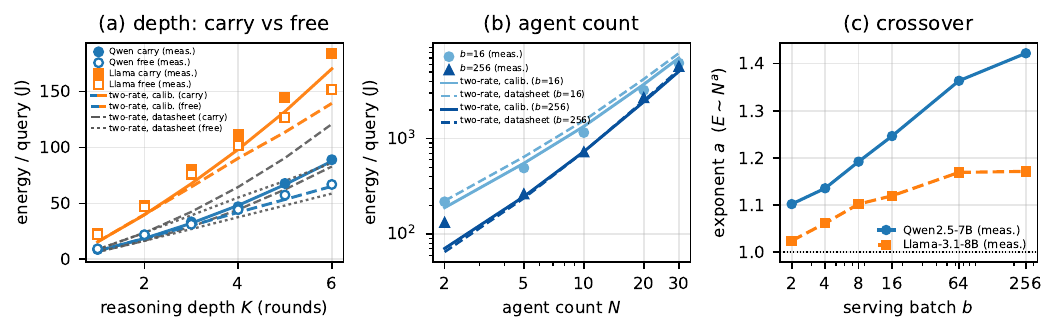}
    \caption{ Measured scaling (A100, vLLM) and model overlays. (a) Reasoning depth on Qwen2.5-7B and
  Llama-3.1-8B: per-query energy vs.\
  rounds, history-carrying vs.\ history-free. (b) Agent count on Qwen2.5-7B: energy vs.\ $N$ at a
  small and a large serving batch. (c) The energy-vs-agent-count exponent $a$ rises with batch on
  both Qwen2.5-7B and Llama-3.1-8B: the prefill/decode crossover.}

  \label{fig:scaling}
\end{figure*}

We rely on the decode from memory bandwidth rather than peak FLOP/s because nameplate
thermal design power overestimates measured energy by up to $4.1\times$: decode
is memory-bandwidth-bound, occupies $77$--$91\%$ of inference time and is nearly
insensitive to compute clock~\cite{mlenergy2025,decode_membound2025}.
Eq.~\ref{eq:cdec} therefore carries a bandwidth term and a batch divisor in place
of a utilization factor, and serving choices (batch size, quantization,
KV-cache reuse) act on energy by influencing  $c_{\mathrm{dec}}(b)$: batching alone a
$3$--$5\times$ reduction~\cite{mlenergy_v3}, latency-constrained Pareto
configurations a further $44\%$~\cite{mlenergy2025} and hence the crossover the
two-rate model makes explicit. The bound presumes saturated batched serving,
which early multi-GPU deployments that ran each device far below its power
budget did not reach: the $3$--$4$\,J per output token reported for a 65B model
in 2023~\cite{samsi2023words} splits the model across $8$--$32$ GPUs that each
draw well under a third of board TDP, and Eq.~\ref{eq:cdec} estimates  such a run
several-fold low.

\begin{figure*}[t]
  \centering
  \includegraphics[width=0.8\textwidth]{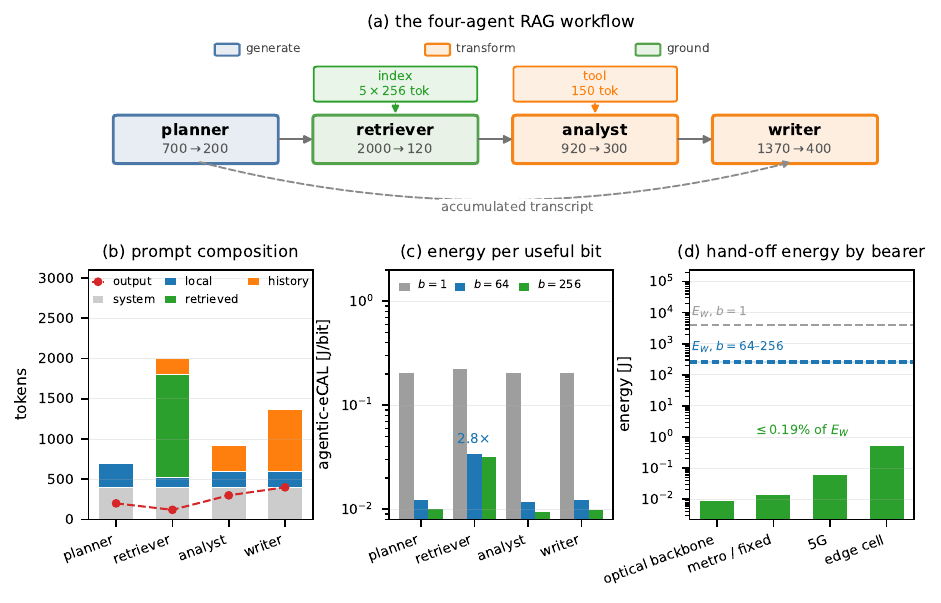}
  \caption{ Four-agent RAG workflow case study (Llama-3 8B, A100). (a)~Sequential execution graph with per-step token counts, tinted by functional class (Sec.~\ref{sec:metric}). (b)~Prompt token composition: transcript accumulation supplies $56\%$ of the writer's prompt. (c)~\agecal{} per agent across serving batches $b \in \{1, 64, 256\}$. (d)~Inter-agent transmission energy across physical bearers (Table~\ref{tab:placement}) vs.\ workflow compute energy $E_W$, demonstrating that transport energy is orders of magnitude below compute.}
  \label{fig:workflow}
\end{figure*}

\subsection{Experimental Validation of Scaling the Call Energy Model} 

Due to the relatively large size of the models corroborated with the context, $E_{call}$ dominates $E_W$. Sec. \ref{sec:callseq} models this process through Eqs. \ref{eq:tokgrow} and \ref{eq:callscale}. Under the same experimental conditions from Sec. \ref{sec:call}, we  validate the scaling.  Fig.~\ref{fig:scaling} shows the measured scaling against reasoning depth (carry
vs.\ free), agent count, and serving batch. We also provide model overlays in Fig.~\ref{fig:scaling} (a) and (b): Eq.~\ref{eq:tworate} evaluated on the same measured token workload at the same serving batch, with i) the calibrated coefficients of Fig.~\ref{fig:validation} and ii) the datasheet numbers. 

In Fig.~\ref{fig:scaling}a, we isolate the cost of conversational memory by comparing history-carrying against history-free loops across reasoning depths $K \in [1, 6]$. In a history-carrying loop, each sequential round re-reads the accumulated transcript of prior steps, driving quadratic prompt token growth in Eq. \ref{eq:tokgrow}. Consequently, measured energy compounds super-linearly with depth (filled markers), whereas history-free execution scales strictly linearly (open markers). At $K=6$, history carry inflates energy by $32.8\%$ on Qwen2.5-7B ($89$\,J vs.\ $67$\,J) and $21.7\%$ on Llama-3.1-8B ($185$\,J vs.\ $152$\,J). Llama consumes roughly $2\times$ the energy of Qwen across all depths due to its higher decode floor ($c_0 = 0.06$ vs.\ $0.02$\,J/token) and larger per-token key-value context footprint ($\gamma = 131$ vs.\ $56$\,kB/token). 

It can be seen from Fig.~\ref{fig:scaling} that  the two-rate model follows the curves given by the measured points: the fitted version closer than the datasheet version. In Fig.~\ref{fig:scaling}a, the hardware datasheet derivation falls $34\%$ below Llama-3.1-8B at $K=6$ because Eq.~\ref{eq:cdec} models pure memory traffic and omits the empirical decode floor ($c_0 = 0.060$\,J/token, driven by PagedAttention fragmentation and idle draw), an omission magnified by Llama's more decode-weighted workload ($p_{\mathrm{in}}\!:\!p_{\mathrm{out}} \approx 2.7:1$ vs.\ $4.5:1$). This is the same observation discussed in Sec. \ref{sec:singlecall} related to Table \ref{tab:validation}.

Fig.~\ref{fig:scaling}b evaluates team width scaling ($N \in [2, 30]$) on Qwen2.5-7B across serving batches $b=16$ and $b=256$. Under all-to-all communication, inter-agent messages scale as $\Theta(N^2)$. At small batch ($b=16$), memory-bandwidth-bound decode is poorly amortized ($c_{\mathrm{dec}} \approx 0.22$\,J/token), dampening the scaling slope ($a \approx 1.25$). Conversely, at production batch ($b=256$), weight-streaming overhead collapses toward the asymptotic key-value floor ($c_{\mathrm{dec}} \to c_0 \approx 0.02$\,J/token),  exposing the compute-bound quadratic prefill term and steepening the scaling exponent to $a \approx 1.42$. Crucially, while batching reduces energy by $38.1\%$ at $N=2$ ($130$\,J vs.\ $210$\,J), this efficiency margin shrinks to just $15.4\%$ at $N=30$ ($4{,}400$\,J vs.
  \ $5{,}200$\,J), demonstrating that batch amortization cannot overcome the quadratic prefill explosion of large, densely coupled multi- agent teams.

Fig.~\ref{fig:scaling}c shows the energy-vs-agent-count exponent rising with batch
accordingly, from $a\!\approx\!1.10$ to $1.42$ for Qwen ($b\!:\!2\!\to\!256$) and from $1.02$ to
$1.17$ for Llama, a direct GPU-energy measurement that confirms the predicted super-linear,
batch-dependent crossover. The milder Llama slope reflects the crossover directly: a
higher decode floor ($c_{\mathrm{dec}}\!\approx\!0.06$ vs ${\sim}0.02$ at $b\!=\!256$) keeps it partly
decode-bound, so architecture sets how far the workflow crosses. The measured exponent 
reflects the partial prefill dominance at the tested batches. The multi-agent \emph{width} axis is
identical: under all-to-all communication each of $N$ agents reads the other $N\!-\!1$, so per-call
input is $\Theta(N)$ and $\sum_k p_{\mathrm{in}}^{(k)}=\Theta(N^2)$, giving the same
$1\!\le\!a\!\le\!2$ law in $N$. Sparser coupling lowers the exponent, and \emph{independent} agents
(no peer context) collapse it to a linear $\Theta(N)$ (the efficient fan-out case). A single crossover
thus governs both reasoning depth ($K$) and team width ($N$). In an agentic workflow, the embodied energy still atomizes with workflow invocations as shown with \ecal{}'s however history  \emph{accumulates} within one workflow.

\subsection{Energy of a RAG Workflow} \label{sec:ragworkflow}

Assuming an agentic workflow including tool call and retrieval, this section aims to study an example 4 agent workflow. Fig.~\ref{fig:workflow}a defines the workflow used as the case study throughout this
section. A planner, a retriever, an analyst and a writer execute in sequence over a shared
transcript, which each step re-reads in full; the retriever adds $1{,}280$ tokens of retrieved
context and the analyst one tool observation of $150$ tokens. Under these assumptions, the four steps consume $4{,}990$ prompt tokens against $1{,}020$ generated, a ratio of
$4.9$. The imbalance is structural rather than incidental: prompts accumulate across steps while
generated output does not, so the transcript supplies $56\%$ of the final step's prompt against
none of the first.

\begin{table}[t]
\caption{Inter-agent transmission as a share of workflow energy, by placement and bearer.}
\label{tab:placement}
\centering
\small
\begin{tabular}{@{}lrrr@{}}
\toprule
Bearer ($\varepsilon$ [J/b]) & co-located & $1$ hop & $3$ hops \\
\midrule
Optical backbone ($10^{-8}$)~\cite{vanheddeghem2012}   & $0$ & $<0.001$\% & $<0.001$\% \\
Metro / fixed ($10^{-7}$)~\cite{lorincz2025ftth}       & $0$ & $<0.001$\% & $<0.001$\% \\
5G RAN ($10^{-6}$)~\cite{xu2020operational}            & $0$ & $0.0001$\% & $0.002$\% \\
Loaded edge cell ($10^{-5}$)~\cite{xu2020operational}  & $0$ & $0.002$\%  & $0.020$\% \\
\bottomrule
\multicolumn{4}{@{}l}{\footnotesize $\varepsilon_{uv}$ is link power divided by achieved throughput. The backbone row is a}\\
\multicolumn{4}{@{}l}{\footnotesize reference value, the others measured; the edge row is~\cite{xu2020operational} outside its good } \\
\multicolumn{4}{@{}l}{\footnotesize coverage window.} \\

\end{tabular}
\end{table}

The case-study workflow is as follows. Four agents run in sequence, the retriever drawing $5$ chunks of $256$ tokens from the index and the analyst invoking one tool that returns $150$ observation tokens; each box gives the step's prompt and generated token counts. Token budgets are stipulated; the values in  Fig.~\ref{fig:workflow} (b) to (d) follows from these through the model of Sec.~\ref{sec:metric}. As can be seen in Fig.~\ref{fig:workflow} b, all agents ingest a system prompt (with grey) and local prompts. The retriever's prompt includes retrieval and history tokens while being dominated by retrieved context. The accumulated transcript grows monotonically across planner, retriever and analyst, and supplies $56\%$ of the writer's prompt, which is what  Eq.~\ref{eq:tokgrow} characterizes. With red dots on the figure, the token outputs of the agents are depicted. It can be seen that they are relatively small compared to the prompts and reach the size of the system prompt only for the writer in this case study.
  
Fig.~\ref{fig:workflow}c evaluates each agent of the workflow with the proposed \agecal from  Eq.~\ref{eq:metric}. It can be seen that the
per-agent figures are nearly uniform at $b=1$, spanning $0.203$ to $0.225$\,J/bit, because decode
dominates and decode energy is proportional to the tokens an agent generates. Batching removes that proportionality and the agents separate: at $b=64$ the planner,
analyst and writer sit between $0.0116$ and $0.0123$\,J/bit whereas the retriever reaches
$0.0340$, a factor of $2.8$, since it prefills $2{,}000$ tokens to emit $120$. The separation
persists at $b=256$, the largest measured batch, where the same three sit between $0.0093$ and
$0.0101$ against the retriever's $0.0317$. The agent that is
least efficient per useful bit is therefore identifiable only in the regime in which the workflow
would actually be served. The workflow as a whole costs $0.0162$\,J/bit against $0.0146$ for its
calls alone, the difference being the tool, the retrieval and the $10\%$ orchestration overhead of
Eq.~\ref{eq:ew}.

Fig.~\ref{fig:workflow}(d) evaluates the energy that  the three hand-offs required by the four-agent workflow of this case study if they were distributed and the token data were transmitted through the bearers of Table~\ref{tab:placement}. The $1{,}290$ tokens crossing the agent
boundaries cost between $8.9$\,mJ on metro fibre and $0.52$\,J on a loaded edge cell, that is
between $0.003\%$ and $0.186\%$ of $E_W$ at $b=64$, and a further $13.9\times$ smaller in
proportional terms at $b=1$, where $E_W$ is correspondingly larger.
Distributing this workflow over any bearer is consequently
free to within the precision of the energy model itself.

\begin{figure}[t]
	\centering
	\includegraphics[width=0.78\columnwidth]{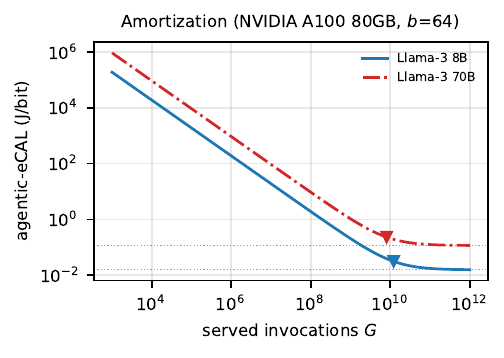}
	\caption{Embodied vs.\ operational \agecal{} as a function of served invocations $G$ for
		Llama-3 8B and Llama-3~70B, on the A100 at $b\!=\!64$; the crossover at
		$G\approx\crossoverG$ for the 8B marks where operation overtakes amortized
		pre-training. Qwen2.5-7B is omitted as it publishes no pre-training
		energy data.}

	\label{fig:amortization}
\end{figure}
\begin{figure*}[t]
	\centering
	\includegraphics[width=0.75\textwidth]{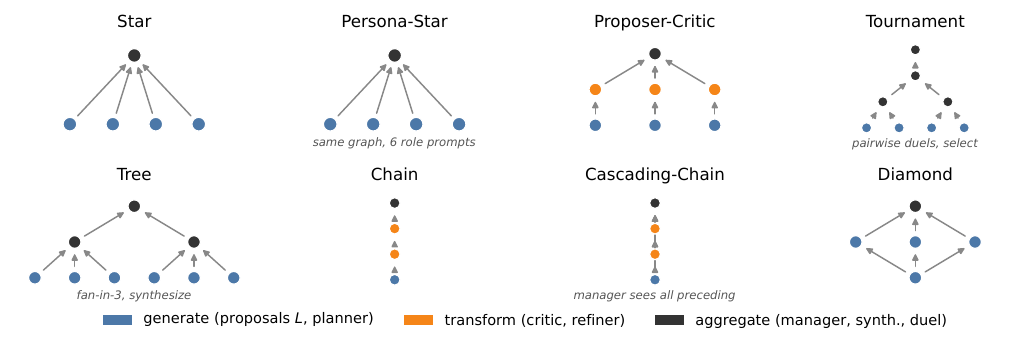}
	\caption{The eight orchestration architectures motivated by \cite{yue2025masrouter,zhang2025maas,yu2026adaptorch} nodes colored by functional class (Sec. \ref{sec:metric}): generate (the $L$ proposals), transform (critics, refiners, synthesizers, duelists), the manager, and Diamond's non-answering planner.  \textsc{Tournament} selects through pairwise duels, \textsc{Tree} merges via fan-in-three synthesis, and \textsc{Persona-Star} shares Star's graph but differs in prompts.  Counts are illustrative.}
	\label{fig:topologies}
\end{figure*}

\begin{figure*}[!t]
	\centering
	\includegraphics[width=0.9\textwidth]{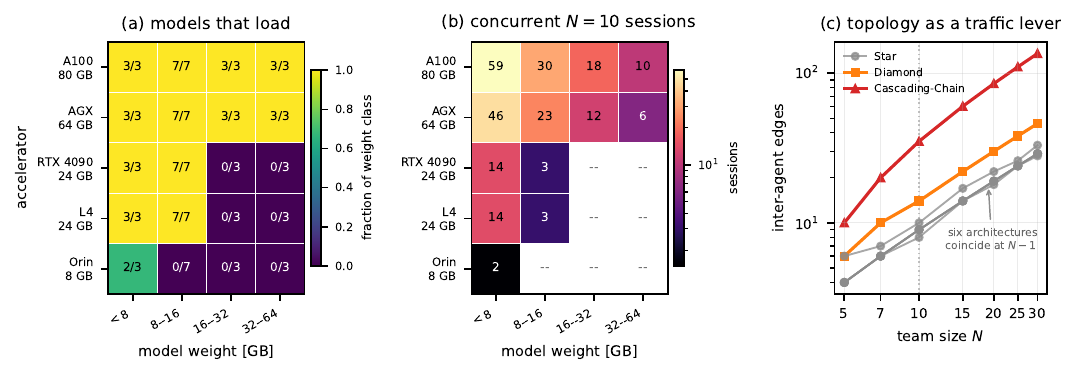}
	\caption{Where an agent team fits across accelerator tiers (Eq.~\ref{eq:c-mem}). (a)~Model loading across 16 open-weight models by weight class ($\beta N_{\mathrm{params}} \le M_u$). (b)~Concurrent serving capacity (sessions of a 10-agent team) under resident KV-cache context ($\gamma \sum_a x_{au} c_a \le M_u - \beta N_{\mathrm{params}}$), showing density is governed by KV width $\gamma$. (c)~Inter-agent edge count $E$ across team sizes $N \in \{5,...30\}$ for the eight orchestration architectures.}
    
	\label{fig:capacity}
\end{figure*}

\subsection{Amortization}\label{sec:amort}

Having validated the operational components of \agecal{} from Eq. \ref{eq:ew}, we now evaluate how the embodied (pre-training) energy amortizes over the total number of workflow invocations, $G$ with results depicted in Fig.~\ref{fig:amortization}. The embodied energy of Llama-3~8B (\embLlamaEightB{}, i.e.\ $1.3$M H100-hours) dominates the per-invocation \agecal{} until $G\approx\crossoverG$. Beyond this point, the fixed pre-training cost atomizes and the operational workflow energy $E_W$ dictates the efficiency floor. 

This workflow invocation amortization is the agentic equivalent of \ecal{}'s single-inference amortization result~\cite{chou2026ecal}. Notably, it exceeds the $2$--$6\times10^{8}$ inference parity threshold reported empirically for the earlier BLOOMz model by a factor of $20$--$60$~\cite{luccioni2024power}. Furthermore, larger models cross this threshold \emph{sooner}: Llama-3~70B reaches operational parity at just $0.66\times$ the invocations of the 8B model. This occurs because the pre-training energy grew by only $4.9\times$ between the 8B and 70B variants, whereas the operational energy per invocation scales directly with parameter count. Consequently, the operational cost per call overtakes the amortized pre-training cost significantly earlier in a massive model's lifecycle.

\section{Placement and Communication Characterization of Agentic Workflows}\label{sec:results-comms}
This section evaluates the placement and communication characteristics of agentic workflows across the edge-cloud continuum, experimentally and numerically grounding the accelerator memory admissibility constraints of Eq.~\ref{eq:c-mem} and the network tier substrate illustrated in Fig.~\ref{fig:overview}. In classical distributed computing, service placement solves an offloading trade-off between local computation and transmission energy. In agentic AI, because inter-agent text transmission is small ($<0.25\%$ of workflow energy), the dominant energy cost of distribution is often not the transmission of inter-agent text itself, but the additional inference and context processing induced by that communication. Consequently, deployment feasibility and serving density across the network are governed principally by the memory capacity boundaries of Eq.~\ref{eq:c-mem}.


\subsection{Where a Team Fits: Accelerators, Models and Transports}\label{sec:capacity}

\begin{figure*}[t]
	\centering
	\includegraphics[width=0.9\textwidth]{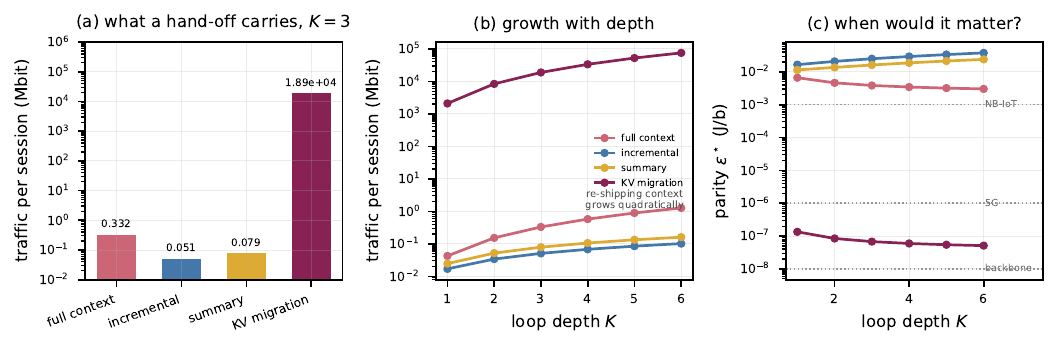}
	\caption{Empirical breakdown of hand-off across network links. (a)~Data volume transmitted per session under four protocols ($K=3$): transcript re-transmission ($0.33$\,Mb), incremental ($0.051$\,Mb), summary ($0.079$\,Mb), and KV-cache migration ($18.9$\,Gb). (b)~Transmitted volume vs.\ loop depth $K$ (quadratic for transcript and cache; linear for incremental/summary). (c)~Parity bearer intensity $\varepsilon^{\star}$ where transport energy equals compute energy: text hand-offs remain orders of magnitude above deployed bearers, while KV-cache migration exceeds compute energy on all cellular bearers.}\label{fig:comm}
\end{figure*}
An agentic workflow is characterised by two properties that a single-model analysis does not expose. The first is heterogeneity of scale: the members of a team need not be, and in practice frequently are not, of the same size, so a site must accommodate whichever class of model the workflow employs. The second is the communication topology, which determines how many messages the members exchange and hence how much traffic a distributed deployment must carry. This subsection treats the two together.

We consider $16$ open-weight models spanning $5.7$ to $50.3$\,GB of weights, grouped into
four classes by the memory those weights occupy. The lightest class, below $8$\,GB, contains
SmolLM3-3B, Qwen2.5-3B and Qwen3-4B. The $8$--$16$\,GB class is the most populous with seven
members, comprising the $7$--$8$B dense models Mistral-7B, OLMo-2-7B, Qwen2.5-7B, Marin-8B,
Llama-3.1-8B and Qwen3-8B together with the mixture-of-experts GPT-OSS-20B, whose routed weights
occupy $11.7$\,GB. The $16$--$32$\,GB class holds the $14$B models Phi-4, Qwen3-14B and
R1-Distill-Qwen-14B, and the heaviest class, $32$ to $64$\,GB, holds Llama-3.3-70B and Qwen2.5-72B
at four-bit weights together with Qwen3.6-27B. Weight class is not, however, a proxy for what a
device can host: quantisation places two $70$B-class models in the same class as a $27$B model at
native precision, and the key-value width within a single class varies by an order of magnitude,
from $36$ to $512$\,kB per token of context. The experiments evaluated $16$ models in a full mesh communication topology on H100 GPU on 6 tasks over $K=3$ communication rounds and teams of $N\in\{1, .., 30\}$. We sub-sample one round and form the communication topologies from Fig. \ref{fig:topologies} assuming one round of reasoning and $N=10$ agents.

The teams themselves are organised by one of the eight orchestration architectures of
Fig.~\ref{fig:topologies}, which differ in how many first-tier proposals they generate and in how
those proposals are subsequently transformed. Star and Persona-Star aggregate $L=N-1$ proposals at
a single manager and differ only in the prompts assigned to the workers. Proposer-Critic,
Tournament, Tree and Diamond interpose a hierarchical or filtering tier, pairing proposals with
critics, resolving them through a pairwise bracket, merging them by fan-in-three synthesis, or
conditioning solvers on separately generated plans, so that $L$ falls to between $5$ and $7$ at
$N=10$ calls. Chain and Cascading-Chain reduce $L$ to unity and advance a single evolving solution
through successive refiners, the latter permitting each refiner to read up to five preceding
reports. Because every edge of these graphs carries one agent's output into another's context, the
architecture fixes the number of inter-agent messages, and with it the traffic a distributed
deployment must transport.

Fig.~\ref{fig:capacity} evaluates the consequences of both properties, taking $N=10$ as a
mid-range team size.  Fig.~\ref{fig:capacity}a evaluates the static weight-loading term
($\beta N_{params} \leq M_{u}$): an $8$\,GB module admits two models of the lightest
class, a $24$\,GB accelerator admits that class and the $8$--$16$\,GB class entire, and $64$\,GB or
more admits all four classes, including the quantised $70$B models. Above the far edge, therefore,
the weight class of a model does not determine whether it can be deployed.

Fig.~\ref{fig:capacity}b reports what the same accelerators sustain once each agent's resident context is
included, expressed in the unit an operator provisions: concurrent sessions of a ten-agent team,
each device holding one copy of the weights together with one context per agent. Capacity ranges
from one or two sessions on the $8$\,GB module to $59$ on the A100, and its ordering does not follow
the weight classes of panel~(a). Within the $8$--$16$\,GB class alone it varies $16\times$, because
capacity is governed by the key-value width $\gamma$ rather than by the weights: GPT-OSS-20B and
Qwen2.5-7B, at $48$ and $56$\,kB per token, sustain $112$ and $87$ sessions, whereas OLMo-2-7B, a
lighter model that retains multi-head attention at $512$\,kB per token, sustains seven. The binding
constraint on serving density is accordingly the context an agent must retain, not the model it
runs.

Fig.~\ref{fig:capacity}c reports how the architectures differ in the traffic they generate. Since each edge
of the orchestration graph contributes one peer message to some agent's prompt, the summed prefill
of a workflow is $N p_{0}+E\tau$ for $E$ inter-agent edges: the shape of the graph does not affect
cost, and only its edge count does. The consequence is that most of the architectures are
indistinguishable. Star, Persona-Star, Proposer-Critic, Chain and, to within one edge, Tournament
and Tree all require $E\approx N-1$ at every team size, so at $N=10$ they lie between $8$ and $10$
edges. Only two separate: Diamond at $E\approx 1.5N$, because each solver is conditioned on a
separately generated plan, and Cascading-Chain at $E\approx 5N$, because each of its refiners
re-reads up to five preceding reports. The dispersion is moreover bounded, widening from $2.5\times$
at $N=5$ to $4.8\times$ at $N=30$ rather than growing with team size.

Summarizing Fig.~\ref{fig:capacity},  we find that the feasibility of deploying an agentic workflow is governed by memory and, within a weight class, by key-value width. The orchestration architecture is a traffic lever only through its edge count, and
for six of the eight architectures that count is the same. What distinguishes the two remaining designs is the re-reading of context, which is the same mechanism that makes prefill the dominant
term of Sec.~\ref{sec:callseq}, and it is therefore in the retention of context rather than in the choice of graph that the design freedom lies.

\begin{figure*}[!t]
  \centering\includegraphics[width=0.9\textwidth]{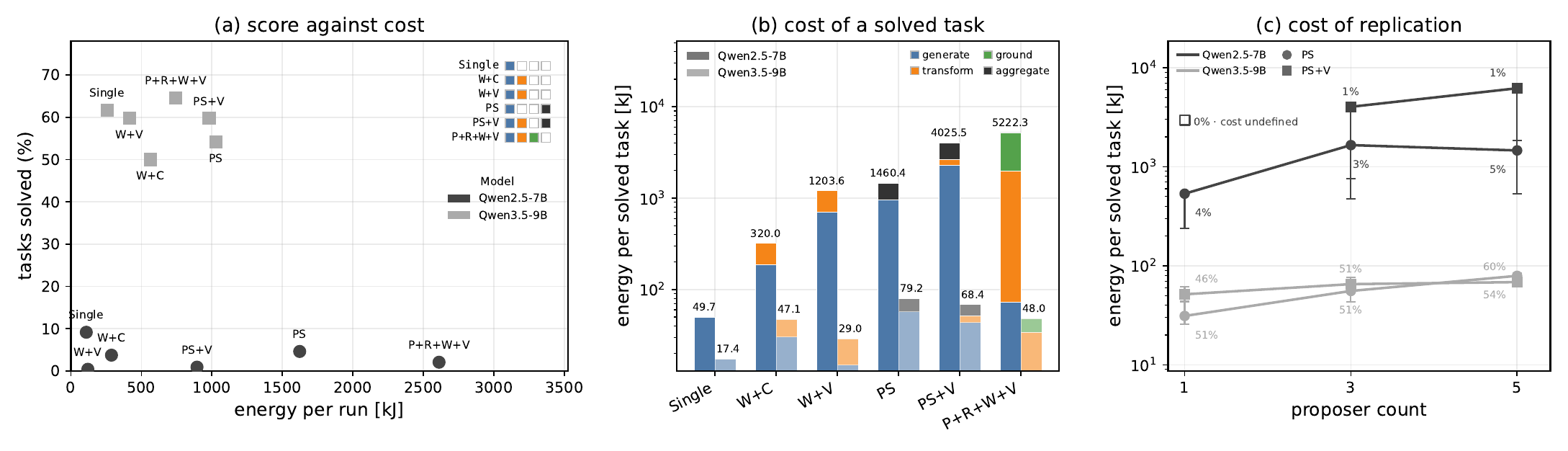}
  \caption{
  Energy and task completion on an infrastructure-repair benchmark for six agent topologies on Qwen2.5-7B and Qwen3.5-9B. Energies are obtained from the measured tokens using the two rate model calibrated for the respective models. (a) Fraction of the twenty-four tasks solved against energy per run. 
  (b) Energy cost per solved task. (c) Energy per solved task against proposer count, annotated with the score attained.
  }
  \label{fig:infra}
\end{figure*}

\subsection{Inter-agent Message Payload Assessment}\label{sec:distributed}
Sec.~\ref{sec:capacity} establishes how many messages an agent team exchanges, while this section studies what the  messages
contain subsampling from the same agent trace dataset. We evaluate four candidate hand-off mechanisms in Fig.~\ref{fig:comm}a for the four-agent workflow at
$K=3$ rounds. Re-shipping the accumulated transcript, as a stateless endpoint must, transmits
$0.33$\,Mbit per session. Carrying only the increment produced since the recipient last observed
the conversation transmits $0.051$\,Mbit, a factor of $6.5$ less, and a bounded running summary
lies between the two at $0.079$\,Mbit. Migrating the key-value cache rather than reconstructing it
transmits $18.9$\,Gbit, five orders of magnitude above the transcript, because the cache entry for
a single token occupies $131$\,kB against the $17$ bits of the token itself. The protocol choice therefore
spans a dynamic range of $3.7\times10^{5}$, compared to $4.8\times$ for the choice of architecture
(Fig.~\ref{fig:capacity}c) and roughly $10^{3}$ across physical network bearers (Table~\ref{tab:placement}):
what a hand-off contains dominates both where it is sent and how it is carried.

These hand-off mechanisms also differ fundamentally in their asymptotic scaling with reasoning depth, as evaluated in Fig.~\ref{fig:comm}b. Re-shipping the transcript grows quadratically in $K$, since each round re-transmits
everything produced before it, rising $30\times$ for  $K=\{1,...6\}$. In contrast, incremental
hand-off grows linearly and rises $6\times$ over the same range, while the bounded summary rises $6.5\times$.
The gap between the stateless and incremental strategies widens with the depth of the loop, reflecting the same history accumulation that makes prompt prefill the
dominant term in Sec.~\ref{sec:callseq}. Cache migration inherits both this quadratic growth and the
five-order magnitude offset simultaneously.

To evaluate under what conditions transmission energy could challenge compute dominance, Fig.~\ref{fig:comm}c inverts the comparison and calculates the parity bearer intensity $\varepsilon^{\star}$ required for transmission energy to equal the accompanying compute energy. For every text-carrying protocol, $\varepsilon^{\star}$ exceeds $3\times10^{-3}$\,J/b across
all reasoning depths,  which represents more than two orders of magnitude above even a loaded edge cell and five orders of magnitude above an optical core: no deployed bearer brings a transcript hand-off within reach of the compute it
serves. Migrating the key-value cache reverses this conclusion entirely: its parity threshold $\varepsilon^{\star}$ falls to
$5.1\times10^{-8}$\,J/b by $K=6$, leaving a margin of only $5\times$ over an optical backbone and
falling below mobile bearers, where cache transport exceeds local compute energy by a factor of twenty. Attention
state is therefore effectively non-transportable across network links, binding an agent's context to the local accelerator that
built it and confirming that the memory capacity boundaries of Sec.~\ref{sec:capacity} cannot be relieved by offloading raw cache tensors over the network.

In this section we showed  that communication efficiency is obtained by transmitting increments rather than transcripts, keeping the cache local. The bearer and the orchestration graph are, by comparison, second-order choices.

\section{Energy per Solved Task on an Infrastructure Benchmark}\label{sec:infra}
To evaluate agentic efficiency on delivered network utility, this section measures task completion on an infrastructure incident benchmark representative of modern Cloud-Native Network Functions (CNFs) at the telco edge.

In 5G-Advanced and emerging 6G architectures, edge computing platforms and O-RAN cloud infrastructures (O-Cloud) rely on containerized network microservices orchestrated via Kubernetes (e.g., k3s for far-edge appliances). Following the  ETSI Zero-touch network and Service Management (ZSM) framework, autonomous agent teams serve as closed-loop diagnostic and remediation engines. We evaluate five multi-agent topologies and a single-agent baseline on a balanced 24-incident subset of the \emph{kubernetes-core} suite of Kubeply's Infra-Bench\footnote{Kubeply, “infra-bench: Open benchmark tasks for evaluating AI
agents on real infrastructure work,” Apr. 2026. 
https://github.com/kubeply/infra-bench}
over ephemeral k3s edge clusters, with eight tasks per difficulty tier. The configurations comprise a single ReAct agent (Single), a worker with critic feedback (W+C), a worker with constraint verification (W+V), parallel proposers followed by a synthesiser (PS), PS with verification (PS+V), and a planner--reviewer--worker--verifier pipeline (P+R+W+V). The benchmark spans incidents like service routing, access control faults, and gives agents 60 minutes and 50 topology turns to diagnose and repair each live cluster before deterministic state verification.

The results depicted in Fig.~\ref{fig:infra} are averaged over at least three runs per configuration, with energy obtained from measured token counts assuming an effective decode batch of $b=64$ using model-specific two rate coefficients: $c_{\mathrm{pre}}=0.023$\,J/token and $c_{\mathrm{dec}}(64)=0.07$\,J/token for Qwen2.5-7B, and $c_{\mathrm{pre}}=0.024$\,J/token and $c_{\mathrm{dec}}(64)=0.15$\,J/token for Qwen3.5-9B.
From Fig.~\ref{fig:infra}a it can be seen that for Qwen3.5-9B, the topologies consume between $257.0$ and $1029.2$\,kJ per run, a factor of $4.0$, while performsnce scores range from $50.0\%$ to $64.6\%$. Single solves $61.7\%$ of tasks; only P+R+W+V exceeds this score, reaching $64.6\%$ while consuming $2.9\times$ more energy. For Qwen2.5-7B, Single solves $9.2\%$ of tasks, while the multi-agent topologies score between $0.4\%$ and $4.6\%$ despite consuming up to $2611.1$\,kJ per run. Fig.~\ref{fig:infra}b reports aggregate energy consumption divided by successful task completions. For Qwen3.5-9B, Single costs $17.4$\,kJ per successful repair, compared with $29.0$ to $79.2$\,kJ for the multi-agent topologies. For Qwen2.5-7B, the corresponding costs are $49.7$\,kJ for Single and $320.0$ to $5222.3$\,kJ for the multi-agent topologies, as several configurations complete fewer than one task per run on average. Fig.~\ref{fig:infra}c examines scaling from one to five proposers. For Qwen3.5-9B, PS cost rises from $31.2$ to $79.2$\,kJ per solved task while its score remains approximately constant. For PS+V, the score rises from $45.8\%$ to $59.7\%$ as the cost increases from $51.5$ to $68.4$\,kJ. Qwen2.5-7B remains below $5\%$ throughout. Replication increases energy in every series, while the largest observed score gain occurs for Qwen3.5-9B PS+V.

\section{Discussion and Limitations}\label{sec:discussion}
While \agecal{} establishes a closed-form foundation for multi-agent lifecycle accounting, three operational dimensions warrant further refinement. First, our formulation treats prompt prefill as computing over the full prefix, which represents a conservative upper bound. In production serving stacks with prefix caching (e.g., vLLM automatic prefix caching and RadixAttention), reusable system prompts and shared conversational branches avoid redundant tensor evaluations, cutting time-to-first-token by up to $4\times$~\cite{ragcache2024}. Incorporating cache hit-rate statistics directly into the prefill summation is a natural extension. Second, for Mixture-of-Experts architectures, the parameter count $N_{\mathrm{params}}$ must be replaced by active routed parameters $N_{\mathrm{act}}$, as unrouted expert weights reside in memory but do not incur per-token matmul FLOPs~\cite{mlenergy_v3}. Third, our empirical calibration captures PagedAttention memory fragmentation on A100; newer microarchitectures featuring hardware-accelerated FP8 or FP4 tensor cores and unified memory will shift the decode roofline slope $a_{\mathrm{dec}}/b$ and reduce the asymptotic floor $c_0$. Finally, embodied pre-training energy is amortized across an estimated global invocation volume $G$, which remains challenging to track for open-weight models redistributed across autonomous private network operators.

\section{Conclusion and Future Work}\label{sec:conclusion}
This paper extended eCAL to agentic AI workflows, introducing agentic-eCAL to account for LLM inference, context accumulation, tool execution, retrieval, inter-agent communication, and embodied energy. The results show that context processing, rather than text transmission, is often the dominant energy cost of multi-agent execution. Under the evaluated network conditions, text-based inter-agent communication contributes only a small fraction of workflow energy, while accumulated context can drive super-linear energy growth. In contrast, KV-cache migration introduces orders-of-magnitude larger payloads and can make network transfer energetically impractical. The infrastructure benchmark further shows that increasing agent count does not necessarily improve task success and can substantially increase energy per solved task.
These findings suggest that energy-aware agentic systems should prioritize context management, incremental hand-offs, local KV state, and agent selection based on measurable task benefit. Future work will incorporate prefix-cache reuse, additional serving stacks and accelerators, speculative decoding, and direct energy measurements of complete agentic workloads.

\bibliographystyle{IEEEtran}
\bibliography{refs}

\end{document}